\PassOptionsToPackage{dvipsnames,svgnames}{xcolor}

\documentclass[preprint,journal]{vgtc}            

\onlineid{1590}
\preprinttext{To appear in IEEE Transactions on Visualization and Computer Graphics.}

\vgtccategory{Research}

\title{EncQA: Benchmarking Vision-Language Models on Visual Encodings for Charts}

\author{%
  \authororcid{Kushin Mukherjee}{0000-0001-5013-6983},
  \authororcid{Donghao Ren}{0000-0001-8666-7241},
  \authororcid{Dominik Moritz}{0000-0002-3110-1053}, and
  \authororcid{Yannick Assogba}{0000-0002-6646-0961}
}

\authorfooter{
  \item
  	Kushin Mukherjee is with Stanford University, work done while at Apple.
  	E-mail: kushinm@stanford.edu.
  \item
  	Donghao Ren is with Apple.
  	E-mail: donghao@apple.com.
  \item
  	Dominik Moritz is with Apple.
  	E-mail: domoritz@apple.com.
  \item
  	 Yannick Assogba is with Apple.
  	 E-mail: yassogba@apple.com.
}

\abstract{%
Multimodal vision-language models (VLMs) continue to achieve ever-improving scores on chart understanding benchmarks. Yet, we find that this progress does not fully capture the breadth of visual reasoning capabilities essential for interpreting charts.
We introduce \eqa{}, a novel benchmark informed by the visualization literature, designed to provide systematic coverage of visual encodings and analytic tasks that are crucial for chart understanding.
\eqa{} provides 2,076 synthetic question-answer pairs, enabling balanced coverage of six visual encoding channels \textit{(position, length, area, color quantitative, color nominal, and shape)} and eight tasks \textit{(find extrema, retrieve value, find anomaly, filter values, compute derived value exact, compute derived value relative, correlate values, and correlate values relative)}. 
Our evaluation of 9 state-of-the-art VLMs reveals that performance varies significantly across encodings within the same task, as well as across tasks. 
Contrary to expectations, we observe that performance does not improve with model size for many task-encoding pairs. 
Our results suggest that advancing chart understanding requires targeted strategies addressing specific visual reasoning gaps, rather than solely scaling up model or dataset size.

}

\keywords{Visual encodings, visualization understanding tasks, machine chart understanding, vision-language models, model benchmarking}

\teaser{
  \centering
  \includegraphics[height=11.5cm, alt={A table of sample chart images from the EncQA benchmarks across 6 tasks and encodings.}]{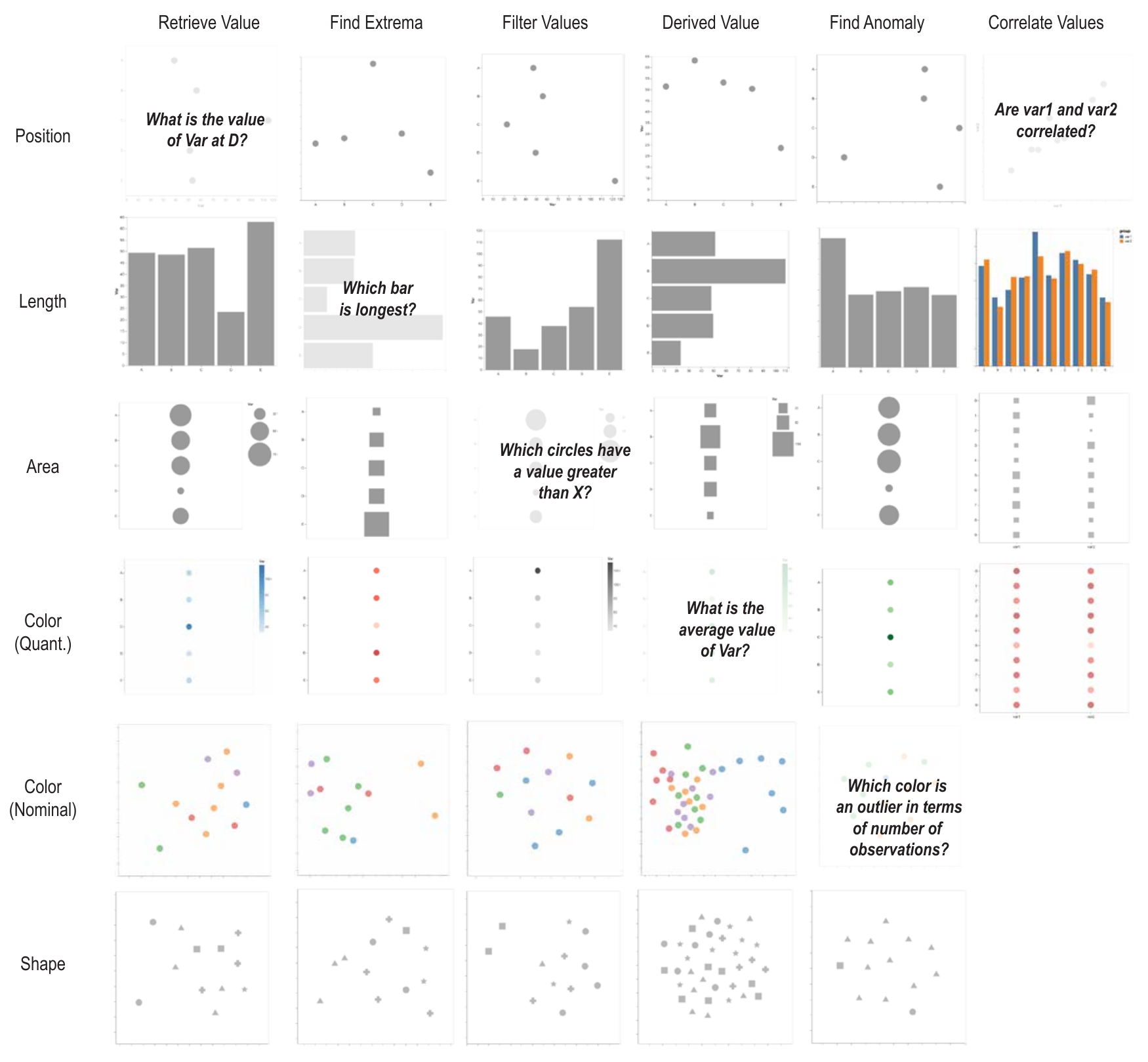}
  \caption{%
  	\eqa{} charts vary across tasks and encodings. Associated questions focus on the visual mapping used to encode the data, for example in the \task{find anomaly} task with the \encoding{area} encoding: ``Which circle is an outlier relative to the rest in terms of area?''
  }
  \label{fig:teaser}
}

\graphicspath{{encqa-tvcg-latex-2024/figs/}{figs/}{figures/}{pictures/}{images/}{./}} 

\usepackage{tabu}                      
\usepackage{booktabs}                  
\usepackage{lipsum}                    
\usepackage{mwe}      
\usepackage{booktabs}
\usepackage{mathptmx}                  

\usepackage{makecell}
\usepackage{multirow}
\usepackage{booktabs}
\usepackage{wrapfig}
\usepackage[dvipsnames]{xcolor}
\usepackage{placeins}
\usepackage{pdflscape}

\newcommand{\eqa}{\textsc{\textbf{EncQA}}}

\newcommand{\task}[1]{\textsc{#1}}
\newcommand{\encoding}[1]{\textsc{#1}}

\newcommand{\varopts}[1]{{\color{Sepia}\{\texttt{#1}\}}}
\newcommand{\varoptsnum}[1]{{\color{Sepia}\{#1\}}}

\begin{document}


\maketitle
\section{Introduction}
Chart understanding has emerged as a key target for multimodal AI systems with frontier generative model releases \cite{openai_gpt-4_2024, gemini_team_gemini_2024,anthropic_introducing_2024} being accompanied by at least one benchmark score on a chart-focused dataset \cite{masry_chartqa_2022, yue_mmmu_2024, huang_pixels_2024}.
The complex nature of chart understanding --- requiring visual perception, abstraction, and reasoning --- makes it a natural candidate to test the limits of models.
Beyond serving as a benchmarking task to assess multimodal reasoning, chart understanding is an ability worth improving in AI systems for practical purposes.
Just as data visualizations themselves are tools that help us communicate information by leveraging visual features to efficiently encode vast data tables, so too can AI systems be thought of as tools that help researchers better understand, interact with, and generate visualizations that effectively communicate intended messages \cite{wu_ai4vis_2022, vaithilingam_dynavis_2024}.
A fundamental condition preceding such use of AI systems for visualization understanding is that their ability to reason about charts should be human-aligned.
That is, models should be able to \textit{decode} information that has been \textit{encoded} using common visual channels --- position, length, color, area, and shape. 
These channels have been shown to be effective for communicating data in visual formats \cite{bertin_graphics_1981, cleveland_graphical_1984}. 
We refer to this ability to effectively use the visual channels and encodings present in charts to make inferences about the information in the chart as \textit{visual reasoning}.

While present chart understanding benchmarks consist of a mix of synthetically generated and web-scraped charts annotated with human-made questions, none of these datasets are created in a manner that allows us to fully isolate the \textit{visual reasoning} component of chart understanding, especially with regard to the use of encoding channels vs. text annotations in charts.
And while researchers have taken aim at generating datasets for testing visual reasoning in a targeted manner \cite{johnson_clevr_2017, liu_mmc_2024, yue_mmmu_2024, kamoi_visonlyqa_2024}, they often do not contain the kinds of visual encodings and tasks identified in the visualization literature as critical for data visualization purposes \cite{bertin_graphics_1981, munzner_visualization_2014}.

In this paper, we address this critical gap by introducing \eqa{}, a benchmark tailored to evaluate chart understanding under a variety of visual encodings and tasks drawn from the visualization literature.
Our contributions are as follows: 
\begin{itemize}
    \item We develop a data-generation framework for creating visualizations with accompanying question-answer pairs spanning six visual encoding channels (e.g., position, area, color) and eight tasks (e.g., retrieve values, compute derived values, find anomalies).
    \item Using this framework, we generate \eqa{} --- a set of 2,250 questions for 2,076 charts. We open source our benchmark items and generation framework at our benchmark website \footnote{Code and dataset are available at https://github.com/apple/ml-encqa}.
    \item We evaluate \eqa{} on a set of 9 vision-language models, including proprietary and open models, and show that model performance on tasks varies substantially by visual encoding used.
\end{itemize}
\section{Background \& Related Work}
\subsection{Visual Encodings for Chart Understanding}

Foundational visualization literature established rankings of visual encoding channels based on their efficacy for different tasks, guiding visualization practice and pedagogy \cite{cleveland_graphical_1984, mackinlay_applying_1988, bertin_graphics_1981, munzner_visualization_2014, wongsuphasawat_voyager_2016, moritz_formalizing_2019, zeng_too_2024}. For instance, Cleveland and McGill \cite{cleveland_graphical_1984} highlighted position as the best encoding for precise ratio judgments. 
However, the same ranking of encodings might not apply for all tasks \cite{bertini_why_2021} demonstrated by Albers et al.'s \cite{albers_task-driven_2014} finding that color-based encodings are better for aggregating estimates (e.g., interpreting time-series data). 
This aligns with theories of perceptual averaging where encoding channels like color allow for efficient aggregate estimation from vision alone without needing explicit computation \cite{correll_comparing_2012}.

We emphasize that visual encodings are distinct from chart types, which result from decisions regarding which visual channel(s) are used to encode the data.

\subsection{Chart Understanding Tasks}

A critique of research surrounding the relative efficacy of visual encoding channels has been that these studies fail to fully characterize the gamut of chart understanding tasks that are natural and common in day-to-day practice \cite{mccoleman_rethinking_2022, bertini_why_2021}.
Several taxonomies have been proposed to systematically characterize visualization tasks \cite{brehmer_multi-level_2013, shneiderman_eyes_2003, wehrend_problem-oriented_1990, amar_low-level_nodate, saket_task-based_2019, quadri_survey_2022}. Wehrend and Clayton \cite{wehrend_problem-oriented_1990} describe tasks in terms of cognitive operations applied to data objects (e.g., identifying, categorizing, ranking, correlating), whereas Shneiderman \cite{shneiderman_eyes_2003} emphasizes interaction-based tasks (e.g., overview, zoom, filter, details-on-demand). 
In contrast, Amar et al. \cite{amar_low-level_nodate} introduce a taxonomy derived from common analytic questions posed by users, focusing on low-level analytic tasks. Their taxonomy comprises ten tasks: \textit{retrieve value, filter, compute derived value, find extremum, sort, determine range, characterize distribution, find anomalies, cluster, and correlate}. In this work, we adopt Amar et al.’s taxonomy because it spans tasks requiring both direct visual extractions---requiring no axes or annotations---and more intricate visual estimates involving averages or correlation judgments, making them ideal for probing visual encoding use in models. This taxonomy has been widely adopted in the visualization literature.

\subsection{Vision Language Models}

Vision-language models (VLMs) generate open-vocabulary text conditioned on visual inputs and textual prompts. These systems combine the rich representations learned by vision encoder models with the expressivity of a powerful large language model (LLM) in order to solve complex visual reasoning tasks (see \cite{huang_pixels_2024} for a more in depth overview).

While chart-specific training improves VLM performance on chart understanding \cite{liu_matcha_2023, masry_unichart_2023, lee_pix2struct_2023, zhou_enhanced_2023, masry_chartgemma_2024}, large general-purpose frontier models also show strong zero-shot capabilities \cite{openai_gpt-4_2024, gemini_team_gemini_2024} especially when prompts are optimized for chart understanding \cite{wu_chartinsights_2024}.
While some works convert charts to non-visual representations such as data tables \cite{liu_deplot_2023}, visualization grammars \cite{bursztyn_representing_nodate, kim_answering_2020} like \texttt{Vega-Lite} \cite{satyanarayan_vega-lite_2017},  or SVG specifications \cite{xu_exploring_2024}, our evaluation focuses on general-purpose VLMs without additional chart-specific annotations or OCR data extraction, targeting their \textit{visual} reasoning capabilities.

\subsection{Large Scale Datasets for Chart Understanding}

Progress in machine chart understanding has been closely tied to the development of large-scale question-answering datasets, both synthetic and web-scraped \cite{methani_plotqa_2020,masry_chartqa_2022,kafle_dvqa_2018, kahou_figureqa_2018}. Early synthetic datasets such as DVQA \cite{kafle_dvqa_2018} and FigureQA \cite{kahou_figureqa_2018} introduced foundational QA tasks over bar and line charts, while later efforts like PlotQA \cite{methani_plotqa_2020} and ChartQA \cite{masry_chartqa_2022} leveraged web-sourced charts for greater variety and complexity. Multimodal benchmarks such as MMMU \cite{yue_mmmu_2024} and MMC \cite{liu_mmc_2024} have further expanded coverage, but their chart-related questions often require domain-specific knowledge, making it difficult to disentangle visual reasoning from subject expertise.

Recent datasets have sought to address these limitations in different ways. CharXiv \cite{wang_charxiv_2024} focuses on real-world charts with questions specifically designed to be answerable from the chart content alone, minimizing reliance on domain knowledge, ChartBench \cite{xu_chartbench_2024} introduces a broader task taxonomy and a wide range of chart types, and ChartInsights \cite{wu_chartinsights_2024} adopts Amar et al.’s \cite{amar_low-level_nodate} taxonomy --- closely aligning with our own—across several chart formats. Notably, Zeng et al. \cite{zeng_advancing_2025} advance chart QA through visualization-referenced instruction tuning, using LLM-based augmentation to increase both the visual diversity of charts and the quality of paired questions, thereby facilitating more robust model training (e.g., via instruction fine-tuning \cite{liu_visual_2023}) and higher quality benchmarks.

Parallel work in visualization and cognitive science has leveraged visualization literacy assessments such as VLAT \cite{lee_vlat_2017} and related studies \cite{verma_evaluating_2024, bendeck_empirical_2024, li_visualization_2024, pandey_benchmarking_nodate, alexander_can_2024} to reveal fine-grained differences between human and VLM performance, as well as to expand the range of visualization QA items using LLMs \cite{cui_promises_2025, zeng_advancing_2025}. While each of these benchmarks expands the diversity, realism, or annotation strategies for chart question answering, they generally evaluate models at the level of overall task or chart type leaving open the question of how models process specific visual encodings.

In contrast, our benchmark is designed for systematic, fine-grained analysis of vision-language model capabilities. By varying visual encoding channels (position, length, area, color, shape) and analytic tasks according to a principled, literature-derived taxonomy, we enable targeted diagnosis of VLM strengths and weaknesses. Our benchmark uniquely combines (1) a focus on visual encodings over chart types, (2) a principled task taxonomy based on established visualization literature, and (3) question design that specifically distinguishes visual reasoning from text extraction or prior knowledge. Previous works have typically incorporated only a subset of these elements, but by systematically targeting all three, our benchmark enables precise analysis of model strengths and weaknesses across visual encodings and analytic tasks.
\section{Motivation}

While there has been steady progress over the past few years on building competent models that can understand visualizations in near human-like ways \cite{huang_pixels_2024, wu_ai4vis_2022}, several empirical and theoretical gaps still remain.
Recent investigations have revealed that many VLMs and vision backbones \cite{bowers_deep_2023, rahmanzadehgervi_vision_2024, fu_blink_2024} \textit{lack basic visual reasoning abilities such as determining whether lines intersect, how shapes are oriented, and counting items in a visual array}.
Such low-level tasks are the building blocks of inference when it comes to visualizations. Thus, to the extent that models are limited in their performance of these tasks, we should also be skeptical about their performance on chart understanding benchmarks.

In addition to these low-level tasks, it is also important to evaluate models on the kinds of naturalistic tasks that we test humans on and design charts to accomplish \cite{verma_evaluating_2024, quadri_survey_2022, amar_low-level_nodate}. While there are many formulations for the process of chart understanding \cite{munzner_visualization_2014, pinker_theory_1990, shah_review_2002}, the process can be broadly captured by 3 distinct components — (1) \textit{`bottom-up' visual perception}, which includes the kinds of visual abilities described above, constitutes the first step and provides the building blocks for visual understanding. (2) \textit{Mapping visual features to data} includes the crucial step of mapping variations in visual features to variations in data. (3) Lastly, an observer must use those mapped values to \textit{answer task-specific questions} \cite{pinker_theory_1990, shah_review_2002}.
An additional factor in the real world is the influence of domain specific knowledge, such as a climate scientist's knowledge of weather patterns influencing their interpretation of a weather map. 
Shah and Hoffner \cite{shah_review_2002} refer to this as `knowledge about content' and an observer's expectations about the data can influence interpretation of visualizations \cite{xiong_curse_2020}.

\begin{figure}[ht!]
    \centering
    \includegraphics[width=0.90\linewidth]{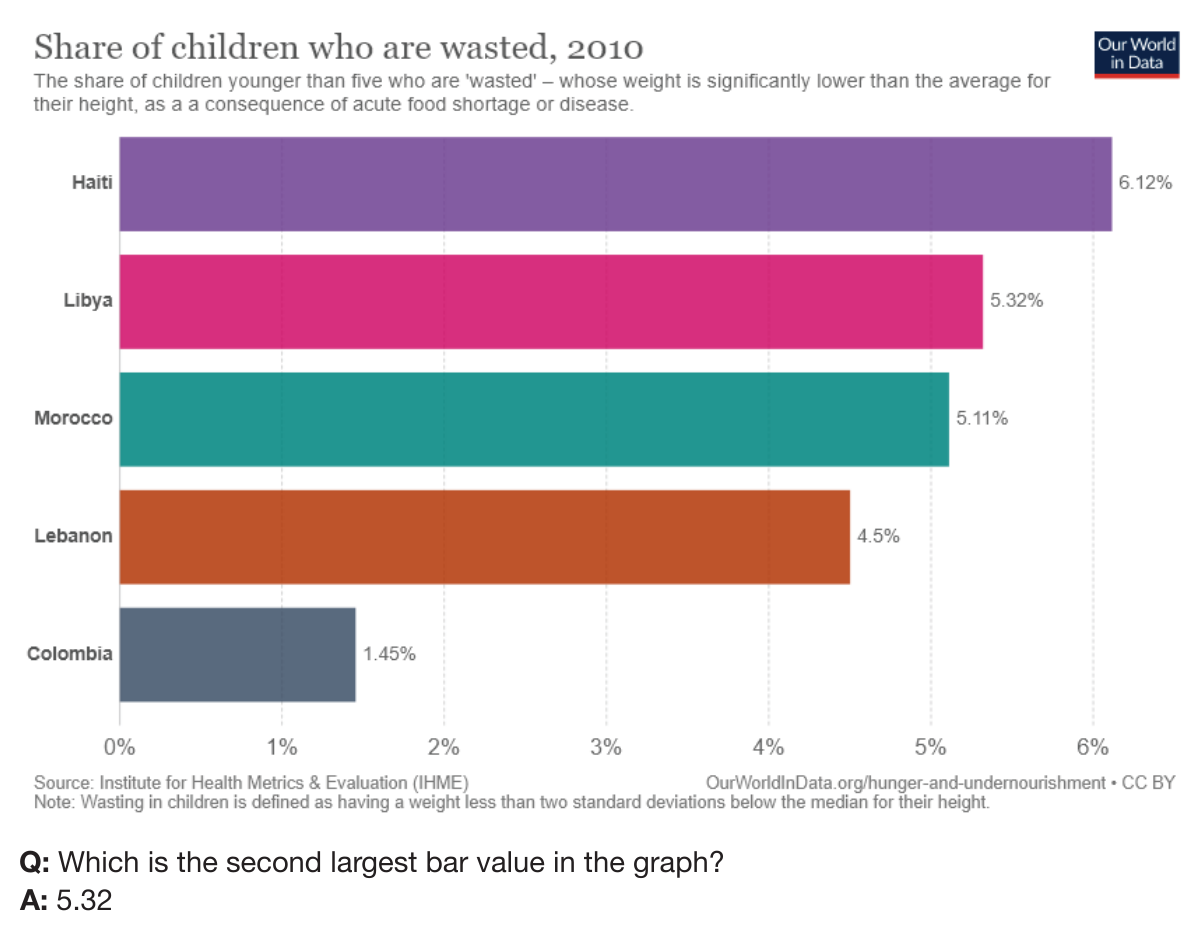}
    \caption{Example Question-Answer pair from ChartQA where the answer could be directly extracted from the text without making a visual judgment of the length of the bar with respect to the axis.}
    \label{fig:chartqa_text_annotation}
    \vspace{-1em}
\end{figure}

In this work, we focus on the main 3 components described above and attempt to minimize reliance on `knowledge about content'.

We design charts and test items that are critically reliant on visual understanding of a chart and cannot be solved using `shortcuts' such as relying on general knowledge of the world, or only using information from the textual annotations (\autoref{fig:chartqa_text_annotation}).

Our benchmark provides a valuable resource for tracking progress on how models are able to decode information from different visual encodings, contextualized by human-relevant tasks. We hope the insights that our benchmark raises can help design data-efficient training strategies to improve VLM chart understanding in a principled manner.

\section{Methods}

\begin{table*}
\centering
\begin{tabular}{@{}lrrrrrrr@{}}
\toprule
\textbf{Task} & \textbf{Length} &  \textbf{Position} &  \textbf{Area} &  \textbf{Color (quantitative)} &  \textbf{Color (nominal)} &   \textbf{Shape} & \textbf{Total}\\ 
\midrule
Retrieve Value &50 &50 & 50& 25& 25 &25 & \textbf{225}\\ 
Find Extrema &100 &100 &100 &50 &50 &50 &\textbf{450} \\
Find Anomaly &100 &100 &100 &50 &50 &50 &\textbf{450} \\
Filter Values  &100 &100 &100 &50 &50 &50 &\textbf{450} \\
Compute Derived Values Exact &50 &50 &50 &25 &50 &50\ &\textbf{275}\\
Compute Derived Values Relative  &50 &50 &50 &25 &25 &25\ &\textbf{225} \\
Correlate Values &100 &50 &100 &50 & N.A. & N.A.\ &\textbf{300} \\
Correlate Values Relative  &50 &25 &50 &25 & N.A. & N.A.\ &\textbf{150}\\
\midrule
&&&&&&\textbf{Total Questions} &\textbf{2,250}
\\
\bottomrule
\end{tabular}
\caption{Number of test items for each task $\times$ encoding cell. We generate at least 25 charts from different underlying datasets and then vary task-encoding specific parameters (e.g., left-right vs. top-down orientation for length encoding charts) to arrive at a final number. See Section \ref{sec:chart_variability} for further details.} 
\label{table:question_numbers}
\end{table*}
\subsection{\eqa{} Benchmark Design}
We introduce \eqa{}, a novel benchmark developed using insights from the visualization and vision science literatures to rigorously evaluate the ability of vision language models to understand data visualizations.
Earlier benchmarks have focused on representing a wide variety of chart types often sourced from the web and have constructed crowd-sourced and researcher-generated QA pairs to create test items \cite{methani_plotqa_2020, tang_vistext_2023,masry_chartqa_2022, masry_chartinstruct_2024 ,wang_charxiv_2024, liu_mmc_2024}.
While this approach allows for the curation of datasets approximately reflecting the kinds of charts used in the real-world, it makes it difficult to ensure an adequate representation of visual encodings \cite{cleveland_graphical_1984, heer_crowdsourcing_2010, mccoleman_rethinking_2022} and tasks \cite{amar_low-level_nodate, quadri_survey_2022, albers_task-driven_2014} that much of visualization research has deemed critical for encapsulating human chart understanding performance.
This raises an important limitation of present benchmarks in that model performance on them might be dominated by successes or failures on specific tasks and encodings. We will explore this limitation in more detail in Section \ref{sec:encqa_chartqa}.
Without a fine-grained understanding of the specific kinds of charts that VLMs struggle with, avenues for progress in this space are limited to scaling benchmark size and chart diversity. 
We posit that this endeavor can be more efficiently directed with a better understanding of the capabilities of VLMs to understand the elemental building blocks of charts. 
Here, we focus on one of these building blocks, namely, visual encodings.

To that end, we structure the questions in the benchmark to target models' understanding of visual channels on visualization-relevant tasks. \eqa{} enables us to evaluate  VLM sensitivity to different visual encodings of data across a range of tasks and contains 2,076 charts paired with 2,250 questions. In the following sections we outline the visual encoding channels and tasks we design \eqa{} around. \\

\subsection{Data Generation}
The underlying values for all chart data are drawn from a normal distribution with a mean at 50 and a standard deviation of 5. 
In cases where one set of data needed to be systematically different from another (e.g., comparing means of two charts), we set the mean of one of the datasets to be 70. For any scatterplot-like charts, we ensured that no two marks were overlapping.

\subsection{Visual Encodings}
Our benchmark tests 6 encoding channels commonly represented in the visualization literature \cite{munzner_visualization_2014, cleveland_graphical_1984, heer_crowdsourcing_2010}---\textbf{position}, \textbf{length}, \textbf{area}, \textbf{color quantitative (lightness)}, \textbf{color nominal (hue)}, and \textbf{shape}.
We use these visual channels to encode two different kinds of variables---\textbf{quantitative} and \textbf{nominal}.
Quantitative variables vary in their numeric values and are encoded using position, length, area, and color lightness.
Nominal variables that represent categorical data are encoded using shape and color. 
We note that there are \textit{no charts that are truly single encoding} charts.
When assigning a test item's visual encoding, we refer to the encoding used to \textbf{encode the data that the question is about}. 
The secondary encoding is often a mapping of the datapoint's category to a position on the axes to prevent data points from overlapping and provides a `named index' to refer to each datapoint.

\subsubsection{Chart Variability}
\label{sec:chart_variability}
When considering these visual encoding channels there are still a number of design choices that visualization designers are left to make. We attempt to capture a number of these in our chart generation process. For each task-encoding pair we generate at least 25 charts, each based on fresh data from our data generation process. In addition to this dataset variability, we apply the following design-specific variants described below:

\textbf{Mark variability}. As the shape of a mark can influence relative area judgments between marks, we generate 25 charts for each of the `circle' and `square' mark types for \encoding{area} encodings. 

\textbf{Orientation variability}. While in human observers, the difference in orientation of bar charts does not lead to drastically different inferences \cite{alallah_oa-graphs_2010}, horizontal and vertical bar charts and dot plots are common variants. Thus, we generate 25 charts for each orientation for \encoding{length} and \encoding{position} and encodings.

\textbf{Color scheme variability}. For \encoding{Color Quantitative} encodings, we cycle through four continuous color schemes defined by \texttt{Vega-Lite}~\cite{satyanarayan_vega-lite_2017}---reds, greens, blues, and grays, within the 25 charts in each task encoding pair. We do not alter the default behavior of Altair \cite{vanderplas_altair_2018} to assign larger data values to the darker end of the color scheme. All charts are on a white background.

Other visual properties of the chart, such as font or mark size, are left to the default values provided by the \texttt{Altair}, the \texttt{Vega-Lite} based python package we use to generate the charts.

\subsection{Tasks}
\label{sec:task_details}

\begin{table*}[ht!]

\begin{small}
\begin{tabular}{@{}>{\raggedright\arraybackslash}p{1.3cm} >{\raggedright\arraybackslash}p{2.2cm} >{\raggedright\arraybackslash}p{2.2cm} >{\raggedright\arraybackslash}p{2.2cm} >{\raggedright\arraybackslash}p{2.2cm} >{\raggedright\arraybackslash}p{2.5cm} >{\raggedright\arraybackslash}p{2.5cm}@{}}
\toprule
\textbf{Task} & \textbf{Length} & \textbf{Position} & \textbf{Area} & \textbf{Color (quantitative)} & \textbf{Color  (nominal)} & \textbf{Shape}\\ 
\midrule
\task{Retrieve value} & What is the value of Var at \varopts{A|B|C|D|E}? & What is the value of Var at \varopts{A|B|C|D|E}? & What is the value of Var at \varopts{A|B|C|D|E}? & What is the value of Var at \varopts{A|B|C|D|E}? & How many \varopts{red|\allowbreak blue\allowbreak|green\allowbreak|orange\allowbreak|purple} circles are present? & How many \varopts{square\allowbreak|cross\allowbreak|triangle\allowbreak|circle\allowbreak|star} shapes are present? \\
\midrule
\task{Find Extrema} & Which bar is the \varopts{longest|shortest}?	 & Which circle is closest to the \varopts{top|bottom\allowbreak|left|right}? & Which \varopts{circle\allowbreak|square} has the \varopts{largest|smallest} area? & Which circle has the \varopts{darkest|lightest} color? & Which color has the \varopts{most|least} observations? & Which shape has the \varopts{most|least} observations? \\
\midrule
\task{Find anomaly} & Which bar is an outlier relative to the rest in terms of length? & Which circle is an outlier relative to the rest in terms of position? & Which \varopts{circle\allowbreak|square} is an outlier relative to the rest in terms of area? & Which circle is an outlier relative to the rest in terms of color lightness? & Which color is an outlier relative to the rest in terms of the number of observations? & Which shape is an outlier relative to the rest in terms of the number of observations? \\
\midrule
\task{Filter values}  & Which bar(s) have a value of Var \varopts{greater |less} than \varoptsnum{$x$}? & Which circle(s) have a value of Var \varopts{greater|less} than \varoptsnum{$x$}? & Which \varopts{circle(s)| square(s)} have a value of Var greater than \varoptsnum{$x$}? & Which circle(s) have a value of Var \varopts{greater|less} than \varoptsnum{$x$}? & Which color(s) have \varopts{more|fewer} than \varoptsnum{$x$} observations? & Which shape(s) have \varopts{more|fewer} than \varoptsnum{$x$} observations?  \\
\midrule
\task{Compute Derived Values Exact} & What is the average value of Var? & What is the average value of Var? & What is the average value of Var? & What is the average value of Var? & What is the average number of observations per color? & What is the average number of observations per shape?\\
\midrule
\task{Compute Derived Values Relative}  & In which chart are the bars longer on average? & In which chart are the circles further to the \varopts{right|top} on average? & In which chart are the areas of the \varopts{circles\allowbreak|squares} larger on average? & In which chart are the circles darker on average? & Which chart has a higher average number of observations per colors? & Which chart has a higher average number of observations per shape? \\
\midrule
\task{Correlate Values}  & Are the lengths of the bars for Var1 and Var2 correlated? & Are Var1 and Var2 correlated? & Are the areas of the \varopts{circles|squares} for Var1 and Var2 correlated? & Are the lightness of the circles for Var1 and Var2 correlated? & N.A. & N.A. \\
\midrule
\task{Correlate Values Relative}  & In which chart are the lengths of the bars for Var1 and Var2 more correlated? & In which chart are Var1 and Var2 more correlated? & In which chart are the areas of the circles for Var1 and Var2 more correlated? & In which chart are the lightness of the cirlces for Var1 and Var2 more correlated? & N.A. & N.A. \\
\bottomrule
\end{tabular}
\end{small}
\caption{Question structure for each task under the different encoding channels. Where possible, the question is expressed in terms of the visual encoding used. Substitution options shown in \varopts{braces}, \varoptsnum{$x$} indicates a numeric substitution.}
\label{table:questions}
\end{table*}

We design our benchmark using six central tasks from  Amar et al. \cite{amar_low-level_nodate}.
The motivation for using this specific task taxonomy is two-fold. First, it is a human behavior-derived granular `low-level' set of tasks that is helpful for assessing fundamental capabilities of VLMs. Second, the widespread adoption of this taxonomy (and variants) in human chart understanding studies facilitates interoperability with existing findings, enabling the identification of discrepancies between model and human capabilities.
Based on existing human perception studies ~\cite{szafir_four_2016, gleicher_perception_2013, rensink_perception_2010, whitney_ensemble_2018} we split the \textit{compute derived values}, and \textit{correlate values} tasks into two variants for a total of eight tasks. These variants ask the model to make a relative judgment between two charts.

Below, we provide brief descriptions of each task:

\begin{itemize}
    \item \task{Retrieve value} is a task focused on retrieving specific values from a chart such as the numerical value corresponding to the height of a bar, or the value implied by a color given a color-ramp legend. This task is presented as a free response question.
    
    \item \task{Find extrema} has to do with finding elements in a visualization with the maximum or minimum value of an encoded variable. For each chart we generate questions to find both the minima and maxima. When generating data for this task, we always ensure there is a single extremal value. This task is presented as a multiple choice question. 

    \item \task{Find anomaly} commonly refers to identifying statistical outliers. Here we focused on outliers in terms of number of observations (for nominal variables), or an observation that has an extreme value of an encoded variable relative to other observations.
    To ensure that the data backing the chart includes an outlier, we began with Tukey's classical definition based on inter-quartile ranges. Specifically, Tukey defines an outlier as an observation that either is less than $Q1 - 1.5\times IQR$ or greater than $Q3 + 1.5 \times IQR$, where $Q1$ and $Q2$ refer to the first and third quartiles and $IQR$ is the inter-quartile range \cite{tukey_exploratory_1977}. 
    In order to make the outlier more visually salient, we instead multiplied the $IQR$ by 3 as opposed to 1.5.
    We randomly designate one of the categories (A - E) as the outlier category and set its value to either $Q1 - 3\times IQR$ (a `small' outlier) or $Q3 + 3\times IQR$ (a `large' outlier). This task is presented as a multiple choice question.

    \item \task{Filter values} refers to the task of identifying marks or observations in a chart that satisfy a particular criterion of being lesser/greater or fewer/more than some reference value. The reference value for the comparison was randomly chosen to either corresponded to the count of the second most or second least frequent category. This task is presented as a free response question and requires the model to produce a list of items.

    \item \task{Correlate Values}
    This task asks the model to estimate if two data series are correlated based on their encodings within a single chart.

    We generate data such that the Pearson correlation coefficient ($r$) was either close to 0.1 or 0.9 \footnote{The correlated dataset was generated by computing the Cholesky decomposition of the desired correlation matrix and computing the product of the decomposed matrix with the first dataset.}.
    This task is presented as a multiple choice question with two options (`no' or `yes').
    
    \item \task{Correlate Values Relative}
    This task asks models to discern \textit{relative} correlation strengths between pairs of datasets. Models are presented with a \textit{pair of charts}, each of which is similar to those shown in the \task{Correlate Values}. One chart in the pair is generated to have $r \approx 0.9$  and the other to have $r \approx 0.1$. The charts are presented side by side and we randomly flip whether the high correlation chart was presented on the right or left. This task is presented as a multiple choice question with two options (`left' or `right').

    Since (Pearson) correlations are reliant on the two sets of data being compared being quantitative, we excluded visualizations of nominal variables from the two correlate values tasks.

    \item \task{Compute Derived Values Exact}. 
    For quantitative variables, this task involves estimating the average value of the variable across multiple categories.
    For nominal variables, the task was to respond with the average number of observations per category, each chart has 5 categories and has either 36 or 6 observations. This task is presented as a free response question and requires the model to produce a numeric answer.
    
    \item \task{Compute Derived Values Relative}. In this task, the objective is to report which of two charts presented has the higher or lower average value of the encoded variable. We generate the data for these charts such that one of the charts always has a clear higher average value than the other. This task is presented as a multiple choice question with two options (`left' or `right').

\end{itemize}

\autoref{table:questions} provides an overview of the question format for each task and encoding channel, while \autoref{fig:teaser} provides a sampling of the charts used in each condition. 

\subsection{Prompt Design}\label{sec:prompt_design}

We construct the text of the final prompt using the following template where each element is separated by a newline:
\textbf{<prefix>
<question>
<suffix>}.

The \textbf{prefix} varies based on the expected answer type for the question. For \textbf{multiple choice} it is \textit{`Answer using only a single word or letter from the options provided.'}, for \textbf{numeric} answers, \textit{`Answer using only a single number.'} and finally for \textbf{list} answers we set the prefix to \textit{`Answer choosing only from the options provided, your answer should be just a simple comma separated list.'}. The benchmark \textbf{question} is then appended followed by an optional \textbf{suffix}. 

For \textbf{numeric} answers the suffix is empty while for \textbf{multiple choice} and \textbf{list} answers the valid options are first randomly shuffled to ensure there is not a positional bias with respect to the correct answer and then appended as a comma-separated list using the template \textbf{`Options: <comma separated list>'}.
The chart image is passed into the model as specified by each model's documentation but always precedes the text described above.

As we added models to our evaluation harness we iterated on prompt design and found this straightforward template to be effective at producing appropriately formatted responses across the models under consideration and provide more detail on that in Section \ref{sec:response_parsing}.

\subsection{Model Evaluation}
We evaluated every model under a greedy decoding scheme in order to make their responses deterministic and set a maximum new token limit of 30 tokens since none of our questions required long responses.

\begin{table}

\begin{small}
\centering
\begin{tabular}{@{}lrr@{}}
\toprule
\textbf{Model} & \textbf{Params} & \textbf{ChartQA} \\
\midrule
GPT-4o & n/a & 85.7 \\
GPT-4o mini & n/a & n/a \\
OpenAI o1 & n/a & n/a \\
\midrule
InternVL2-2B & 2.1B & 76.2 \\
InternVL2-8B & 7.3B & 83.3 \\
InternVL2-26B & 26B & 84.9 \\
\midrule
Molmo-7B-D & 7B & 84.1 \\
Phi-3.5 Vision & 4.2B & 81.8 \\
ChartGemma & 2.9B & 80.2 \\
\bottomrule
\end{tabular}
\caption{Models evaluated, their approximate parameter counts and their reported ChartQA scores.}
\label{tab:model_suite}

\vspace{-2em}
\end{small}
\end{table}

We focus our evaluation on 3 commercial models and 6 open models listed in \autoref{tab:model_suite}. We select these based on their high scores on ChartQA \cite{masry_chartqa_2022}, a well established chart understanding benchmark. We believe these models provide a representative snapshot of the current state of chart understanding models at various sizes at the time of writing. 

\subsection{Response Parsing}
\label{sec:response_parsing}
We implement a simple regular expression based response parsing pipeline to extract answers from the model response text. We test each response against a list of 5 regexes that were written after looking at sample responses from all the models we tested. One separates the answer from leading phrases like ``Answer is:'' and two more extract comma-separated lists for set answers, which are then converted to set objects. Our next regex matches against word representations of numbers such as ``one, two, three, etc.'', and our final one matches numeric digits with an optional decimal part to handle numeric answers. If none of these patterns match, the whole string is considered as the answer. We found that the vast majority of model responses follow the requested format and we are able to successfully parse valid (though not necessarily correct) answers for almost all models. \textit{InternVL2-2B} produces only 4 responses we couldn't parse, and manual inspection confirmed they were incorrect. \textit{ChartGemma} produced 279 of these non-parseable responses, most of which are refusals to perform the task. \textit{Molmo-7B-D} produced 86 non-parseable responses all of which are in the \task{Compute Derived Value} task, as it tries to produce a reasoning chain to answer the question and runs out of tokens before producing an answer.

\subsection{Metrics}

A motivating question was whether \eqa{} would uncover differences in how well VLMs solve chart understanding tasks across different visual encoding channels.
To measure model performance across tasks using a commensurable metric, we use \textit{accuracy}, a simple yet effective metric used broadly in both evaluations of human and VLM data visualization understanding \cite{huang_pixels_2024, verma_evaluating_2024}.
Given the different response formats across tasks (numeric responses, multiple choice selections, list generation), we use the following question type-specific measures of accuracy.
For multiple choice questions, we use \textit{exact match} to compare model responses to true labels.
For numeric responses, we compute the \textit{relaxed accuracy} metric \cite{methani_plotqa_2020} and like other works evaluating chart question answering, we set the threshold for a correct response to be within 5\% of the true label.
For list responses, such as in the the filter value task, we use \textit{set equality} between the generated list and true label.
Because of the relatively large number of responses from ChartGemma that we are not able to parse via regular expression, we further use an LLM-as-judge approach \cite{chiang_can_2023} to evaluate whether ChartGemma responses are equivalent to the ground truth. We use GPT4o as the judge and score an answer as correct if either the regex method or the LLM indicate that the response is correct.
When aggregating scores up to the task or model level, we first \textit{compute mean accuracy within each task-encoding pair} then aggregate those scores to get task-specific and then model-specific scores. This is done in order to account for the different number of stimuli in each task-encoding pair.

\begin{table}[t!]
\setlength\extrarowheight{-1pt}
\setlength{\tabcolsep}{2pt}
\caption{\eqa{} accuracy for 2 leading commercial models and the 3 open source models from our suite with the highest scores on ChartQA. Bold numbers indicate the encoding that each model performs the best on within a task.}
\label{tab:big_table}
\begin{small}
\begin{tabular}{@{}p{2.1cm}p{2.2cm}rrrrr@{}}
\toprule
\textbf{Task} & \textbf{Encoding} & \makecell[c]{GPT \\ 4o} & \makecell[c]{OpenAI \\ o1} & \makecell[c]{Phi \\ 3.5} & \makecell[c]{InternVL2 \\ 26B} & \makecell[c]{Molmo \\ 7B-D} \\
\midrule
\midrule
\multirow{0}{*}{\makecell[l]{Retrieve Value }}& Length & 0.38 & 0.34 & 0.64 & \textbf{0.72} & 0.64 \\
& Position & 0.54 & 0.48 & 0.72 & 0.48 & 0.60 \\
& Area & 0.20 & 0.14 & 0.12 & 0.10 & 0.08 \\
    & Color Quantitative & 0.16 & 0.16 & 0.04 & 0.00 & 0.04 \\
& Color Nominal & \textbf{0.96} & 0.76 & \textbf{0.84} & 0.56 & 0.92 \\
& Shape & \textbf{0.96} & \textbf{0.80} & 0.64 & 0.48 & \textbf{0.88} \\
\midrule
\multirow{0}{*}{\makecell[l]{Filter Values }}& Length & \textbf{0.91} & 0.79 & \textbf{0.59} & \textbf{0.79} & \textbf{0.44} \\
& Position & 0.84 & \textbf{0.91} & 0.33 & 0.41 & 0.26 \\
& Area & 0.79 & 0.62 & 0.12 & 0.34 & 0.05 \\
& Color Quantitative & 0.76 & 0.60 & 0.14 & 0.10 & 0.22 \\
& Color Nominal & 0.26 & 0.52 & 0.32 & 0.08 & 0.08 \\
& Shape & 0.50 & 0.56 & 0.38 & 0.12 & 0.08 \\
\midrule
\multirow{0}{*}{\makecell[l]{Find Extrema }}& Length & 0.99 & 0.97 & \textbf{0.99} & \textbf{1.00} & \textbf{1.00} \\
& Position & 0.93 & 0.97 & 0.93 & 0.83 & 0.98 \\
& Area & 0.96 & 0.98 & 0.88 & 0.85 & 0.97 \\
& Color Quantitative & \textbf{1.00} & \textbf{1.00} & 0.70 & 0.72 & 0.98 \\
& Color Nominal & 0.98 & \textbf{1.00} & 0.72 & 0.66 & 0.50 \\
& Shape & 0.96 & 0.92 & 0.64 & 0.60 & 0.46 \\
\midrule
\multirow{0}{*}{\makecell[l]{Find Anomaly }}& Length & 0.96 & 0.96 & 0.85 & 0.79 & 0.81 \\
& Position & 0.88 & 0.95 & 0.91 & 0.71 & \textbf{0.89} \\
& Area & 0.85 & 0.93 & \textbf{0.96} & \textbf{0.98} & 0.66 \\
& Color Quantitative & \textbf{0.98} & \textbf{0.98} & 0.90 & 0.76 & 0.84 \\
& Color Nominal & 0.50 & 0.56 & 0.50 & 0.50 & 0.24 \\
& Shape & 0.50 & 0.62 & 0.48 & 0.50 & 0.44 \\
\midrule
\multirow{0}{*}{\makecell[l]{Compute Derived \\ Value Exact }}& Length & \textbf{0.40} & 0.32 & \textbf{0.34} & 0.24 & \textbf{0.34} \\
& Position & 0.26 & 0.30 & \textbf{0.34} & 0.22 & 0.32 \\
& Area & 0.16 & 0.10 & 0.16 & \textbf{0.40} & 0.22 \\
& Color Quantitative & 0.16 & 0.16 & 0.08 & 0.12 & 0.20 \\
& Color Nominal & 0.04 & \textbf{0.78} & 0.00 & 0.00 & 0.00 \\
& Shape & 0.00 & 0.56 & 0.00 & 0.00 & 0.00 \\
\midrule
\multirow{0}{*}{\makecell[l]{Compute Derived \\ Value Relative }}& Length & 0.98 & \textbf{1.00} & 0.78 & 0.80 & 0.78 \\
& Position & 0.70 & 0.70 & 0.64 & 0.68 & 0.68 \\
& Area & \textbf{1.00} & 0.98 & \textbf{1.00} & \textbf{1.00} & \textbf{1.00} \\
& Color Quantitative & \textbf{1.00} & \textbf{1.00} & \textbf{1.00} & \textbf{1.00} & \textbf{1.00} \\
& Color Nominal & 0.64 & \textbf{1.00} & 0.44 & 0.44 & 0.28 \\
& Shape & 0.28 & \textbf{1.00} & 0.48 & 0.64 & 0.52 \\
\midrule
\multirow{0}{*}{\makecell[l]{Correlate Values }}& Length & 0.51 & 0.71 & 0.58 & 0.54 & \textbf{0.50} \\
& Position & \textbf{0.96} & \textbf{0.92} & \textbf{0.60} & 0.56 & \textbf{0.50} \\
& Area & 0.75 & 0.86 & 0.50 & \textbf{0.58} & \textbf{0.50} \\
& Color Quantitative & 0.70 & 0.90 & 0.50 & 0.56 & \textbf{0.50} \\
\midrule
\multirow{0}{*}{\makecell[l]{Correlate Values \\ Relative }}& Length & 0.82 & 0.76 & 0.62 & 0.60 & 0.52 \\
& Position & \textbf{1.00} & \textbf{0.96} & \textbf{0.88} & 0.68 & \textbf{0.64} \\
& Area & 0.52 & 0.74 & 0.48 & 0.58 & 0.44 \\
& Color Quantitative & 0.56 & 0.84 & 0.28 & \textbf{0.72} & 0.36 \\
\midrule
\midrule
{\makecell[l]{EncQA Score}}& & 0.67 & 0.73 & 0.55 & 0.54 & 0.51 \\
\bottomrule
\end{tabular}
\end{small}
\vspace{-1em}
\end{table}

\section{Results}

\subsection{VLM Sensitivity to Encoding Channel Varies within and across Tasks}

\begin{figure}[t]
    \centering
    \includegraphics[width=\linewidth]{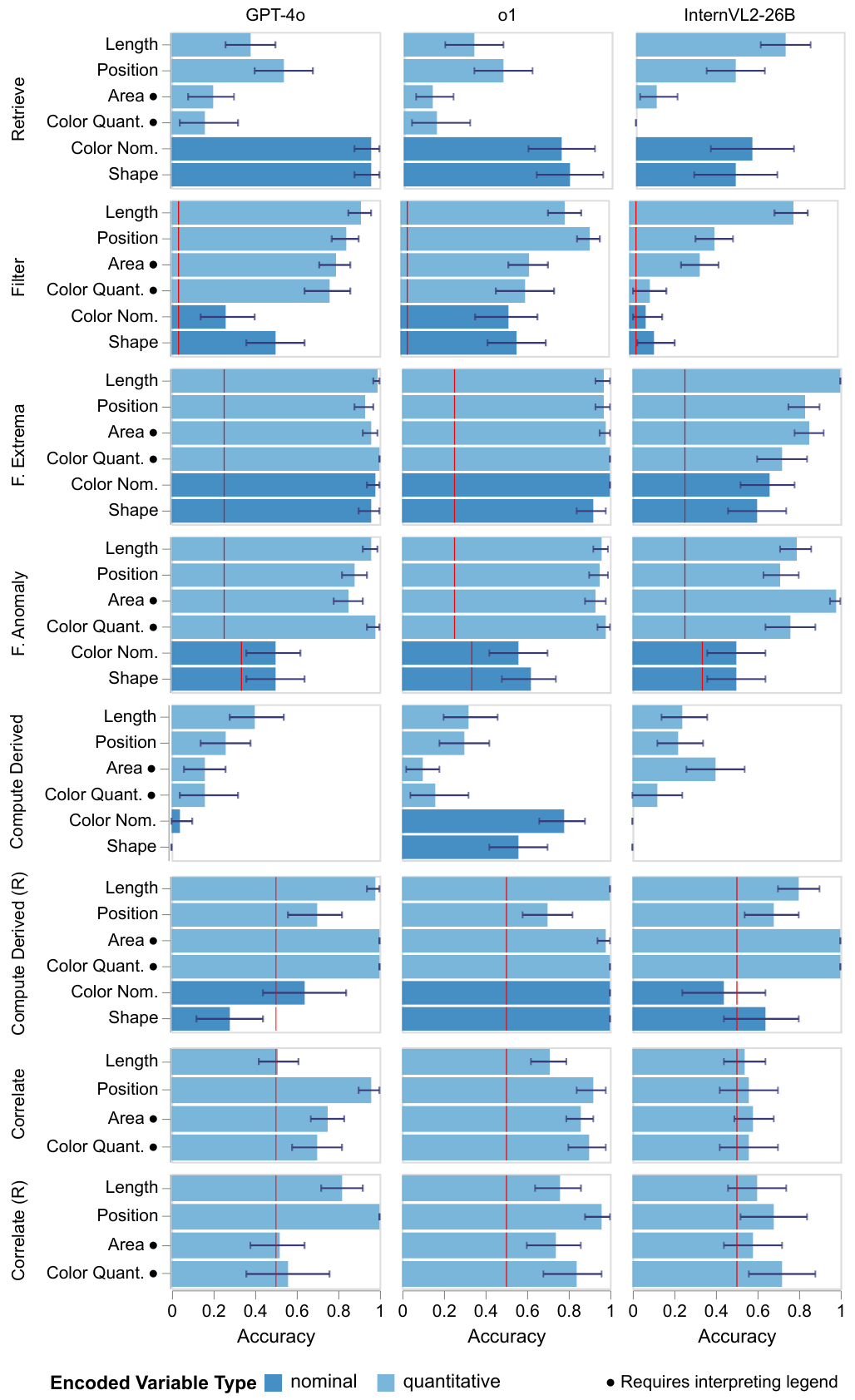}
    \caption{Model accuracy per visual encoding for each task. (R) indicates `Relative' variants of the task. Red lines mark chance level performance for multiple choice questions. Error bars indicate 95\% confidence intervals computed via the bootstrap method}
    \label{fig:encoding_task_accuracy}
    \vspace{-1em}
\end{figure}

Due to space constraints we focus our analysis here on the performance of two commercial `frontier' models, namely, GPT-4o and OpenAI o1, and a top performing open model, InternVL2-26B. We believe these provide a snapshot of the current state-of-the-art for chart understanding. \autoref{fig:encoding_task_accuracy} displays their performance on the EncQA benchmark. \autoref{tab:big_table} contains detailed results for these models as well the other 2 top-performing open models that we tested. We observe a number of interesting phenomena that we describe below.

\textbf{Encodings that require interpreting legends are significantly more difficult than those requiring reading values off of an axis.} In the \task{retrieve value} task, the \encoding{area} and \encoding{color quantitative} encoding charts both require the use of a legend to estimate the relative size or lightness of the target mark. All models do better with length and position encodings (which do not require a legend) for this task (\autoref{fig:encoding_task_accuracy} first row). This effect is less consistently observed in the \task{filter values} task, which doesn't require as precise an estimate of values as \task{retrieve value} (GPT-4o in \autoref{fig:encoding_task_accuracy}, second row).

\textbf{Simple counting tasks are easy, counting and comparing is significantly harder.} In \eqa{}, questions for encodings of nominal data (i.e., \encoding{shape} and \encoding{color nominal}) are framed as counting tasks, reflecting how these encodings are generally used in visualizations (e.g., ``How many blue circles are in the chart?'', ``Which shape has the most observations?''). When these are \textit{just} counting tasks (i.e., \task{retrieve value} and \task{find extrema}) the models perform quite well.
However, when they require extra aggregation or judgment, model performance drops dramatically. For example if asked ``How many colors have more/fewer than x observations'' (\task{Filter Values}), GPT-4o's performance drops to less than 50\% compared to the near 100\% performance seen in the simpler counting tasks. An exception to this is the o1 `reasoning' model which performs substantially better at this task at the cost of significantly more inference time compute. We discuss this more in our exploration of Chain-of-Thought (CoT) prompting which we explore in Section \ref{sec:cot}

\textbf{Models struggle with visual estimates of correlation.} The \task{correlate} tasks ask the model to estimate whether two visual variables are correlated, and we see that in many cases performance is close to random chance. Some notable exceptions are GPT-4o when using position encodings (for which we use a scatterplot --- arguably the most common type of plot for estimating correlations), as well as o1, which performs best overall. We also note that there isn't a consistent change in behavior when moving to the arguably easier `relative' version of the task where the model is asked to judge which of two charts shows data that are more strongly correlated. While GPT-4o's performance drops for two out of four encodings, InternVL2-26b performance remains relatively similar across encodings.

\subsubsection{Favored Task-Encoding Variants}
\label{sec:favored_task_variants}

As detailed in Sections \ref{sec:chart_variability} and \ref{sec:task_details}, we include a number of variants for certain encodings and task formulations. While models do not demonstrate notable differences for chart orientation variants (horizontal or vertical), or mark type (circle or square), or color scheme we do note some variants that reveal marked differences in performance.

\begin{figure}[t]
    \includegraphics[width=0.95\linewidth]{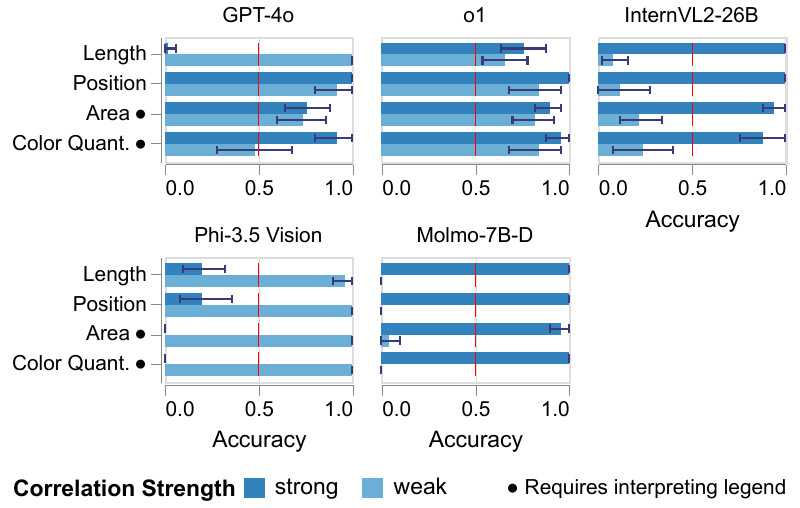}
    \caption{Effect of correlation strength on accuracy for the \task{Correlate Values} task.}
    \label{fig:correlation_strength}
\end{figure}

\autoref{fig:correlation_strength} shows the effect of correlation strength on accuracy for the \task{Correlate Values} task. Across encodings, most models can \textbf{only perform this task if there is a strong correlation}. Phi-3.5 notably reverses this pattern, and GPT-4o displays an exception to this for only one encoding channel (length).

\begin{figure}[t]
    \includegraphics[width=0.95\linewidth]{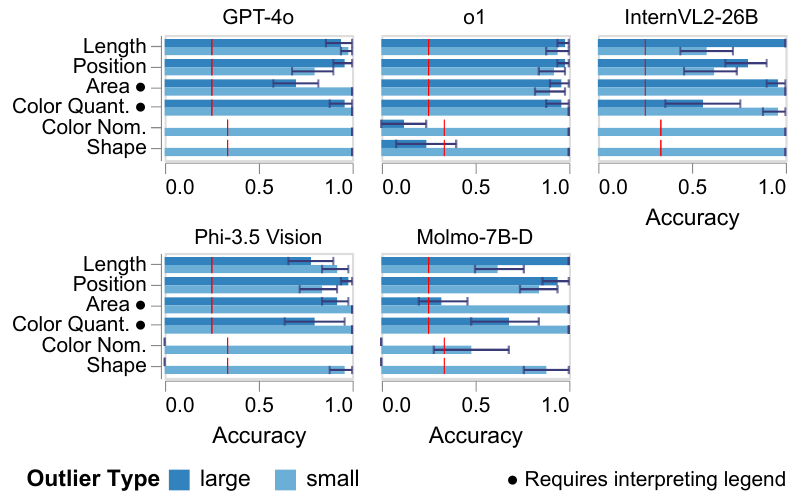}
    \caption{Effect of outlier type on accuracy for the \task{Find Anomaly} task.}
    \label{fig:outlier_type}
\end{figure}

We also observed that in the \task{Find Anomaly} task, when using \encoding{Shape} or \encoding{Color Nominal} encodings (i.e., performing a counting task), overall performance is around 50\%. On further inspection, we find that all these models are able to do the task \textbf{at near 100\% performance when the outlier has the `smaller' count, and near 0\% when the outlier category has more observations} (\autoref{fig:outlier_type}). This could reflect a bias in the models' conception of outliers. We further note that this drop in performance is not improved by zero-shot Chain-of-Thought prompting or when using the o1 reasoning model (Section \ref{sec:cot}).

The synthetic, controlled nature of the \eqa{} is a key enabler for targeted measurement of these fine grained differences. Users of the benchmark can easily generate more chart-question pairs for task-encoding pairs of interest.

\subsection{How far off are numerical responses?}

\begin{figure}[ht!]
    \centering
    \includegraphics[width=1\linewidth]{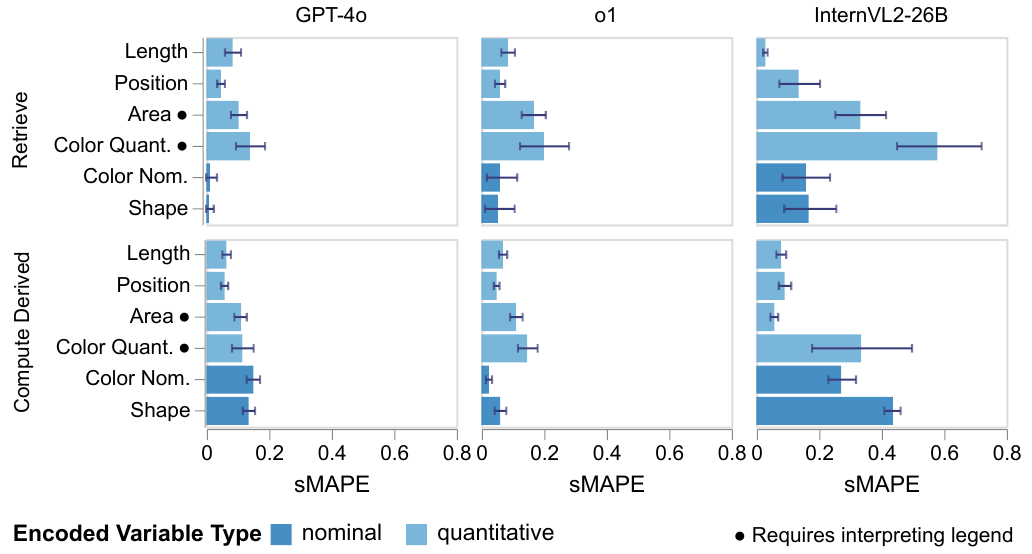}
    \caption{Symmetric Mean Absolute Percentage Error for numeric responses (lower is better).}
    \label{fig:mape}
\end{figure}

Two of our tasks require a precise numerical response, namely \task{Retrieve Values} and \task{Compute Derived Values}. We previously reported relaxed accuracy scores for these tasks (\autoref{fig:encoding_task_accuracy}). While useful, this metric gives limited insight into the \textit{degree} to which answers are correct or incorrect. In order to provide a finer grained look at model responses, we computed the symmetric mean absolute percentage error (sMAPE) between the model predictions and ground truth for all numeric responses (\autoref{fig:mape}). sMAPE is a continuous measure of the distance of a model's predictions from the true values. It is computed as follows: $$sMAPE = \frac{1}{n} \sum_{k=1}^{n} \frac{|T_k - P_k|}{(|T_k| + |P_k|)/2} $$ where $T_k$ and $P_k$ are the true and predicted values respectively of the $k$th question.

For the two encodings that require interpreting a legend (\encoding{area} and \encoding{color quantitative}), performance on \task{Compute Derived Value} is comparable or better than performance on \task{Retrieve Value}, this is somewhat surprising as the latter task is ostensibly a sub-task of the former.

\subsection{Chain-of-Thought/Reasoning Improves Performance For Only a Limited Set of Task-Encoding Pairs}
\label{sec:cot}

\begin{figure}[t!]
    \centering
    \includegraphics[width=1.0\linewidth]{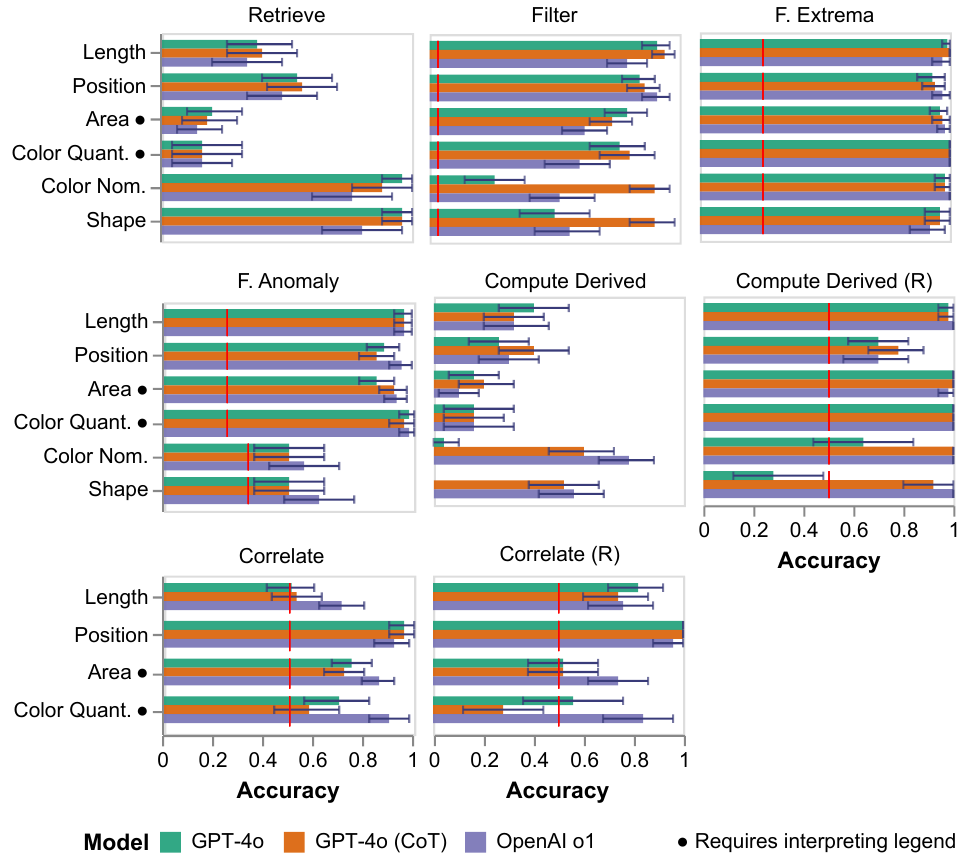}
    \caption{CoT improves performance mostly higher level tasks like compute derived values or correlate and for some counting tasks. However some task-encoding pairs see performance drops with CoT compared to direct prompting.}
    \label{fig:cot_encoding_task_accuracy}
    \vspace{-1em}
\end{figure}

A question that emerges, particularly when considering tasks that require a greater degree of judgment, is whether they benefit from the extra compute steps that zero-shot Chain-of-Thought (CoT) prompting \cite{kojima_large_2024}, or `reasoning' models provide \cite{openai_learning_2024}. Zero-shot CoT asks a model to `think step by step' before producing its final answer. We test this approach using GPT-4o and o1\footnote{For o1 we do not include the `think step by step' instruction as the model is trained specifically to do this and the developer documentation recommends avoiding that addition}. For GPT-4o, we allow the model to generate up to 500 new tokens (rather than 30 in the base experiment), and for o1 we set its reasoning effort parameter to `medium' (the default) and observe it uses anywhere from 400 to 8,590 tokens per question (the final answers are still generally under 10 tokens). Both models provide a structured output response API, making it easy to separate the reasoning chain from the final answer. 

We observe that most task-encoding pairs are not improved by CoT/reasoning (\autoref{fig:cot_encoding_task_accuracy}). We do however, see improvement in three tasks (for certain encodings), namely \task{filter values}, \task{compute derived value} and \task{compute derived value relative} with both CoT and reasoning. These tasks can be seen as requiring a counting step followed by an aggregation step or set of comparisons. 

We do observe some task-encoding pairs where reasoning (o1) outperforms CoT (though they are often comparable), namely in the \task{correlate values} and \task{correlate values relative} tasks, for the \encoding{area} and \encoding{color quantitative} encodings. As observed in \autoref{fig:correlation_strength}, o1 is the only model to consistently handle both correlation strengths across all encodings. Surprisingly, in the \task{filter values} task, we see the reverse for these two encodings, where CoT (and even non-CoT GPT-4o) outperforms the more token intensive reasoning model that is designed to be more capable at visual reasoning\footnote{\url{https://platform.openai.com/docs/guides/reasoning-best-practices\#5-visual-reasoning}}.

\subsection{Model Size Does Not Reliably Correlate with Performance}
\label{sec:model_size}

\begin{figure*}[ht]
    \centering
    \includegraphics[width=\textwidth]{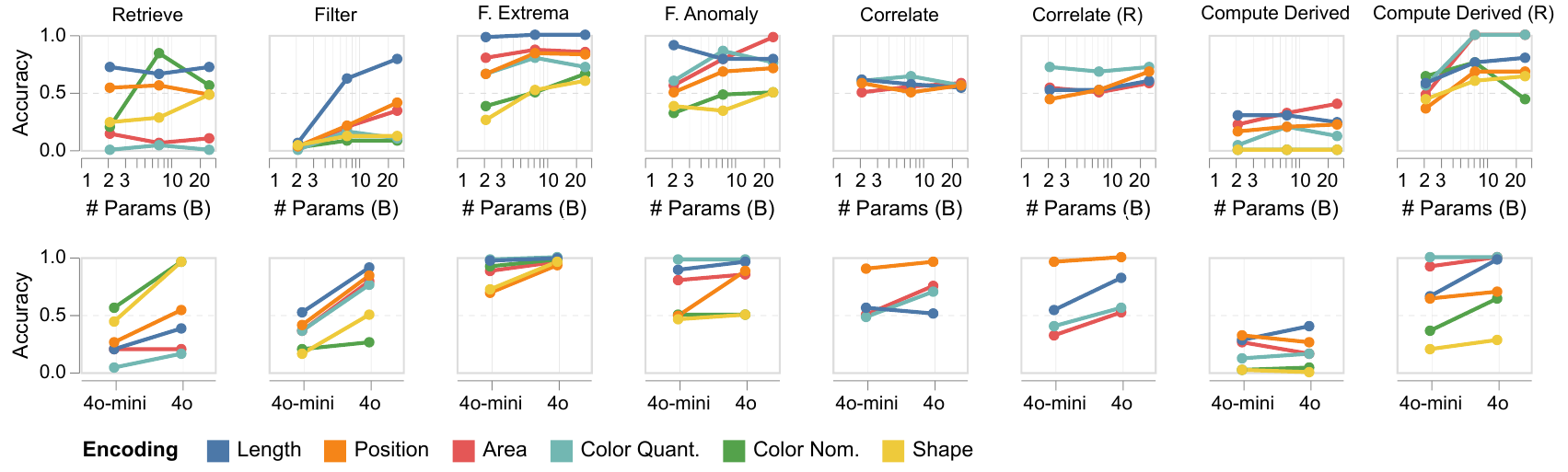}
    \caption{Accuracy vs Model Size for (\textbf{TOP}) three models in the \textbf{InternVL2} Model Family (2B, 8B and 26B, x-axis on a log scale) and (\textbf{BOTTOM}) GPT4o and GPT4o-mini. Some tasks and encodings do see improvements with model scale, but many have no or sub-linear improvement with size. The two smaller InternVL2 models have the same size vision encoder (300M params) while the larger one scales the vision encoder to 6B params.}
    \label{fig:size_internvlm}
\end{figure*}

A useful property of the deep learning revolution has been a steady predictable increase of model performance with scale (in compute or data), where larger models trained on more data outperform smaller models \cite{kaplan_scaling_2020}. However we observe that this is often not the case for task-encoding pairs in \eqa{}. \autoref{fig:size_internvlm} (top) shows results for three sizes of model from the InternVL2 family. Many task-encoding pairs do not improve as the model size increases or only do so sub-linearly. And even when there is improvement in task performance it is rarely across all encoding channels. 

This may suggest that \eqa{} helps highlight elements of capabilities that do not simply scale with model or dataset size.
Improving these capabilities might require more targeted collection of data or novel pretraining objectives.
While we do not have access to the parameter counts of the commercial models, we present a similar (though coarser) analysis, comparing GPT-4o with GPT-4o-mini (\autoref{fig:size_internvlm} (bottom)).

\section{Discussion}

\subsection{What does \eqa{} test that ChartQA doesn't?}
\label{sec:encqa_chartqa}

\begin{figure}[ht!]
    \centering
    \includegraphics[width=0.9\linewidth]{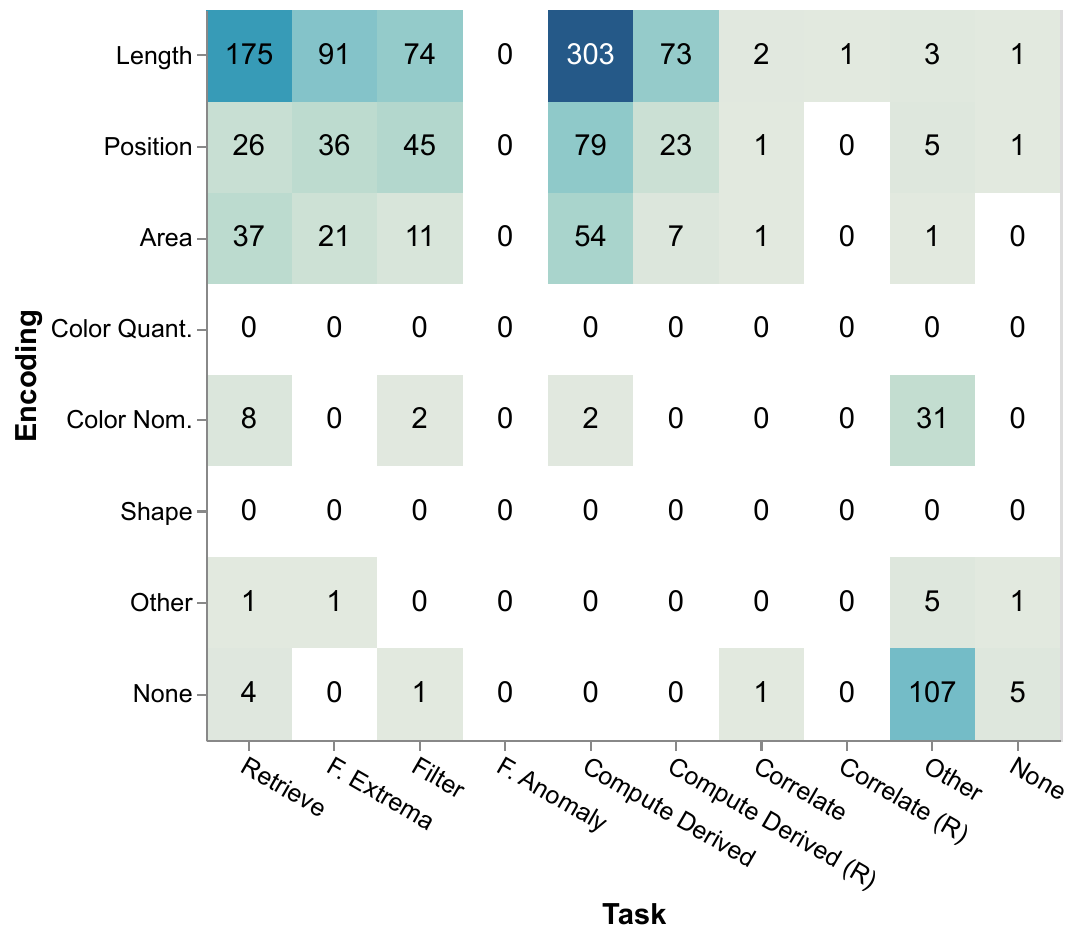}
    \caption{ChartQA test samples categorized by encodings and tasks used in this work. `Other' represents tasks outside of the 8 tasks we considered.}
    \label{fig:chartqa_encoding_task_counts}
    \vspace{-1em}
\end{figure}

A key motivation behind our work is the position that it is difficult to ascertain the extent to which current chart evaluation benchmarks focus on visual reasoning and whether said benchmarks represent the range of  tasks and visual encodings central to visualization understanding.
While there are no doubt other cognitive components to chart understanding beyond just visual reasoning \cite{shah_review_2002, pinker_theory_1990}, visualizations are most helpful when they take advantage of the visual system's ability to easily extract information from charts \cite{franconeri_science_2021}. In order to further ground this position, we annotated the test split of \textbf{ChartQA} \cite{masry_chartqa_2022} --- the most used chart understanding benchmark at the time of writing --- with metadata indicating: the chart understanding task for the question, the primary visual encoding for the data under question, and whether text annotations within the chart image itself would allow for answering the question without the need to actually decode the information using the visual encoding used to create the marks.
We took the \textbf{ChartQA-test human-labeled} split and two of the authors first reviewed approximately 50 diverse question-answer pairs to develop guidelines for how to categorize them; then one of the authors annotated all 1,250 test questions.

We found that most test items (81\%) \textit{could} be answered without visual reasoning over encodings and instead by extracting text from the chart image (e.g. \autoref{fig:chartqa_text_annotation}).
While redundantly encoding information using text and visual features is indeed helpful (e.g., for memorability \cite{bateman_useful_2010}), the presence of text annotations in these charts makes it difficult to differentiate models' text recognition and localization capabilities, from their use of visual encodings and associated axes to solve chart understanding tasks.

\autoref{fig:chartqa_encoding_task_counts} shows the distribution of charts with respect to encoding and task. We observe that most charts in ChartQA test use \encoding{length} or \encoding{position} encodings, followed by \encoding{area} encodings. We note that none of the charts use \encoding{color} for quantitative variables (such as in a heatmap or choropleth map) or \encoding{shape} encodings. 
With regards to task distribution, \task{compute derived value} and \task{retrieve value} are the most represented tasks while \task{find anomaly} and \task{correlate values} have no or little representation respectively.
There were a significant number of ``Other'' tasks represented, but these mostly consisted of tasks like correctly mapping a legend value to its referent (e.g., ``What does Green bar represents?'' in a stacked bar chart where one series is colored green), extracting information from titles (e.g. \textit{``Which country data shown in the Chart?''} in a chart where the title indicates which country the chart is about), or even reading values from axes or annotations (e.g. \textit{``Find missing data of the sequence 24, \_ ,32, 33, 42?''} in a chart where bars are labeled with those numbers). 
There were also cases where the visual encoding could be said to be ``None.'' These were questions about an aspect of the chart that wasn't encoded using any visual feature (e.g. \textit{``Is there a value 30 in the dark blue line?''} in a chart with no axes or tick marks but where each data point is annotated with its value).

This analysis demonstrates the increase in coverage of encodings and tasks that \eqa{} provides.

\section{Limitations}

We acknowledge that \eqa{} has some limitations.
In particular, for a given encoding and task pair, there are multiple valid charts that could be created to satisfy those constraints. We generally chose the simplest design that fit the criteria, and thus the visual complexity of our stimuli is relatively low. 
Nevertheless, we observe that even these simple charts reveal significant performance gaps and differences in response patterns across models. 
Future iterations of \eqa{} could expand the variety of charts used to fill in the encoding-task matrix and the visual complexity of the charts (particularly as model performance saturates for task encoding pairs). The number of encodings and tasks tested could also be increased in future. We also do not test the sensitivity of models to image resolution. 

While we set up \eqa{} to have independent visual encodings for each task type, as noted earlier, charts often use more than one encoding. Even if only a single encoding is relevant for answering a question, a second encoding is often required to prevent the marks from overlapping. To further isolate models' sensitivity to specific visual encodings, even simpler visual stimuli could be developed to test for the fine-grained differences between visual encoding channels (e.g., position vs. length) without the limitation of needing to be like naturalistic visualizations such as in \cite{rahmanzadehgervi_vision_2024, chae_decomposing_2025}.

While we adopted a popular taxonomy of visualization tasks \cite{amar_low-level_nodate} for principled reasons, there exist alternative ways of formalizing the visualization task space \cite{galesic_graph_2011, ge_calvi_2023, boy_principled_2014, lee_task_2006} --- some more high-level than the ones presently considered. 

Lastly, our results on model accuracy vs. model-size is limited to one open-source model and one closed-source model family (for which we have limited insight into the differences between the large and small models), potentially limiting their generalizability.

\section{Conclusion \& Future work}

In this paper, we introduce a new benchmark, \eqa{}, that tests vision-language models on their ability to perform visual reasoning tasks relevant to chart understanding. We address the need for a rigorous benchmark that varies charts in both the visual encoding channel and tasks that are evaluated using these charts, with a targeted focus on visual reasoning as opposed to testing models' general word knowledge. 
As the field considers how AI chart understanding might be effectively advanced we ought to consider which strategies might be most beneficial for both evaluating and improving model capabilities. The dominant strategy we observe in the field is a drive towards larger datasets that have more realistic charts and more complex questions, whether scraped from the internet or generated by advanced VLMs \cite{cui_promises_2025}. In the design of \eqa{} we propose an alternative approach to evaluation that instead starts from principles of visualization design and perception and constructs test items to measure specific perceptual abilities. We consider these two approaches to be complementary and have described aspects of VLM chart understanding behavior that \eqa{} is uniquely able to isolate. 

We find that \eqa{} reveals performance differences in VLMs across visual encoding channels (length, position, area, color, shape) for diverse visualization tasks. Accuracy patterns vary across models and even top-performing models, including those optimized for reasoning, fail to consistently answer questions correctly across all tasks and encodings. We also observe that error patterns cannot be explained as a function of model size. These results underscore the need for combining insights from visualization and machine learning, with rigorous evaluation of models on targeted benchmarks.

While we have focused on evaluation of VLM capabilities, we see room for future work that explores how synthetic datasets and frameworks for parametrically generating benchmarks could be used to \textit{improve} the capabilities of VLMs on real-world chart understanding tasks.

\section{Acknowledgments}
We thank our colleagues at the Visualization team at Apple for helpful feedback and discussion. We also thank our anonymous reviewers for their comments that helped strengthen the work.
\bibliographystyle{abbrv-doi-hyperref}

\bibliography{encqa-paper}
\end{document}